# STARN-GAT: A Multi-Modal Spatio-Temporal Graph Attention Network for Accident Severity Prediction

Nobin
Department of Urban and Regional Planning
Khulna University of Engineering and Technology
Khulna, Bangladesh
nobin5625@gmail.com

Rifat
Department of Urban and Regional Planning
Khulna University of Engineering and Technology
Khulna, Bangladesh
rifat2217058@stud.kuet.ac.bd

*Abstract*— Accurate prediction of traffic accident severity is critical for improving road safety, optimizing emergency response strategies, and informing the design of safer transportation infrastructure. However, existing approaches often struggle to effectively model the intricate interdependencies among spatial, temporal, and contextual variables that govern accident outcomes. In this study, we introduce STARN-GAT, a A Multi-Modal Spatio-Temporal Graph Attention Network, which leverages adaptive graph construction and modality-aware attention mechanisms to capture these complex relationships. Unlike conventional methods, STARN-GAT integrates road network topology, temporal traffic patterns, and environmental context within a unified attention-based framework. The model is evaluated on the Fatality Analysis Reporting System (FARS) dataset, achieving a Macro F1-score of 85.0%, ROC-AUC of 0.91, and recall of 81% for severe incidents. To ensure generalizability within the South Asian context, STARN-GAT is further validated on the ARI-BUET traffic accident dataset, where it attains a Macro F1-score of 0.84, recall of 0.78, and ROC-AUC of 0.89. These results demonstrate the model's effectiveness in identifying high-risk cases and its potential for deployment in real-time, safety-critical traffic management systems. Furthermore, the attention-based architecture enhances interpretability, offering insights into contributing factors and supporting trust in AI-assisted decision-making. Overall, STARN-GAT bridges the gap between advanced graph neural network techniques and practical applications in road safety analytics.

*Keywords—graph neural network, graph attention network, spatio-temporal modeling, traffic safety, multimodal data fusion*

## I. Introduction

The National Highway Traffic Safety Administration [1] has identified accurate prediction of accident severity as essential for developing proactive traffic safety management systems. Predicting traffic accident severity is challenging due to the complex interaction of multiple factors operating at different spatial, temporal, and contextual scales [2]. Traditional approaches to predict severity have predominantly relied on statistical models and basic machine learning techniques [3]-[6] that process features independently, failing to capture complex interdependencies. These basic methods can make predictions but cannot properly capture how transportation networks connect or how accidents involve multiple factors [7].

Graph neural networks have emerged as a powerful paradigm for modeling complex relational systems, demonstrating remarkable success in transportation applications such as traffic flow prediction and network analysis [8], [9]. The natural alignment between road networks and graph structures makes GNNs particularly compelling for traffic safety applications [10]. However, most existing work applies graph neural networks primarily to accident occurrence or frequency prediction rather than severity classification [11]. When severity prediction is addressed, these approaches typically employ basic graph construction methods that rely solely on spatial proximity or simple topological connections, missing the richer connectivity patterns that characterize real road networks [12].

Moreover, while the importance of multi-modal data integration is widely recognized in traffic safety research [13], current approaches lack fusion mechanisms that can effectively combine spatial road network characteristics, temporal patterns, and environmental contextual factors. Existing multi-modal models [14], [15] generally use simple concatenation strategies that treat all modalities equally, failing to capture the dynamic scenarios. This limitation is significant because the importance of spatial, temporal, and contextual factors changes based on different road conditions and situations. [16].

To address these limitations, this paper introduces STARN-GAT (Spatio-Temporal Graph Attention Network), a comprehensive multi-modal graph neural network architecture specifically designed for traffic accident severity prediction. Our approach combines a comprehensive graph construction method that maps road network complexity through multi-criteria connectivity, considering topological relationships, spatial proximity, and functional similarity with adaptive neighborhood definitions. The system utilizes a multi-modal neural architecture that leverages graph attention networks for spatial encoding, specialized temporal networks for pattern extraction, and attention-based fusion mechanisms for data integration.

The primary contributions of this work include:

- A graph construction framework that comprehensively model road network connectivity through multiple criteria and adaptive parameters.
- Application of attention-based multi-modal fusion to graph neural networks for accident severity prediction.
- Extensive experimental validation demonstrating significant performance improvements over existing approaches across multiple evaluation metrics and two real-world datasets.

The remainder of this paper is organized as follows. Section 2 surveys related work. Section 3 formally constitute the problem. Section 4 describes the STARN-GAT architecture in detail, including graph construction, feature engineering, and training protocols. Section 5 outlines our experimental setup; section 6 presents quantitative results and ablation analyses. Finally, section 8 concludes.

## II. RELATED WORK

### A. Traditional Accident Severity Prediction

Early research in traffic accident severity prediction mainly employed statistical methods, with logistic regression and ordered probit models serving as foundational approaches. Abdulhafedh [3] developed multinomial logistic regression models for accident severity classification, achieving moderate performance but failing to capture complex non-linear relationships inherent in traffic safety data. Similarly, Ozbay [4] employed ordered probit models to analyze the relationship between road characteristics and accident severity. The introduction of machine learning techniques marked a significant advancement in predictive performance. Random Forest models gained popularity due to their ability to handle mixed data types and provide feature importance rankings. Sam and Gulia [5] demonstrated the effectiveness of ensemble methods in accident severity prediction, achieving improved accuracy over traditional statistical approaches. Support Vector Machines (SVMs) [6] were also extensively explored.

However, these traditional approaches [3] – [6] share fundamental limitations: they treat accident records as independent instances, failing to capture the spatial correlations between nearby road segments, and they process features in isolation without considering complex dependencies between features.

### B. Deep Learning in Transportation Safety

The application of deep learning techniques to transportation safety problems has gained significant momentum in recent years. Park et al. [17] developed CNN-based models for accident severity prediction using road imagery, achieving promising results but remaining limited to visual features. Recurrent Neural Networks (RNNs) and their variants, particularly Long Short-Term Memory (LSTM) networks, have instituted a new era of prediction in temporal traffic safety modeling. Zhang et al. [18] employed LSTM networks to model temporal dependencies in accident occurrence patterns, demonstrating superior performance over traditional time-series methods.

The emergence of attention mechanisms marked a significant advancement in deep learning approaches to traffic safety. Transformer-based architectures have begun to appear in transportation research, with some studies applying self-attention mechanisms to model complex dependencies in traffic data. However, most existing deep learning approaches [17], [18] in traffic safety remain limited to single modalities and fail to integrate the multi-modal nature of accident data effectively.

### C. Graph Neural Networks in Transportation

Graph Neural Networks have emerged as a powerful paradigm for modeling transportation networks, leveraging the natural graph structure of road systems. Early applications focused primarily on traffic flow prediction, with Graph Convolutional Networks (GCNs) demonstrating remarkable success in capturing spatial dependencies in traffic data. Yu et al. [8] introduced spatiotemporal GCNs for traffic forecasting, establishing the foundation for graph-based transportation modeling. Graph Attention Networks (GATs) further advanced the field by introducing learnable attention mechanisms that could adaptively weight the importance of different neighbors in the graph. Veličković et al. [19] demonstrated that attention-based approaches could capture complex relationships more effectively than fixed convolutional approaches.

In the context of traffic safety, graph-based approaches have shown promise but remain focused primarily on accident occurrence prediction rather than severity classification. Recent work by Guo and Liu. [20] applied basic GCNs to accident prediction, achieving improved performance over traditional methods but employing simplistic graph construction that relies solely on spatial proximity. Similarly, Jin et al. [21] developed graph-based models for accident hotspot identification, demonstrating the potential of graph approaches but lacking multi-modal integration and severity-specific optimization.

### D. Multi-Modal Learning and Feature Fusion

Multi-modal learning has gained significant attention across various domains, with transportation applications beginning to explore the integration of diverse data modalities. Early multi-modal approaches in transportation primarily employed simple concatenation strategies, combining features from different sources without considering their complex interactions. More sophisticated fusion strategies have emerged, including early fusion (feature-level), late fusion (decision-level), and hybrid approaches. However, most existing work in traffic safety continues to rely on basic concatenation methods [14] that treat all modalities equally and fail to capture the dynamic importance of different information sources across varying accident scenarios.

Attention-based fusion mechanisms represent the current state-of-the-art in multi-modal learning [15], enabling models to dynamically weight the importance of different modalities based on their relevance to specific prediction tasks. While these approaches have demonstrated success in other domains, their application to predicting traffic accident severity within graph neural network frameworks remains largely unexplored. This represents a significant opportunity, as the complex relationships between spatial road network characteristics, temporal patterns, and environmental factors in determining accident severity could benefit substantially from attention-based integration strategies [16].

## III. PROBLEM FORMULATION

The traffic accident severity prediction problem is formulated as a multi-class classification task within a spatial-temporal-contextual framework. Given an accident record an occurring at location $l$ = (latitude, longitude) at timestamp **t** with contextual information **c**, our objective is to predict the severity class; **y** ∈ {*no injury*, *minor*, *moderate*, *severe*} with associated confidence scores.

Formally, we define the prediction function as:

$$f: \mathcal{A} \rightarrow \mathcal{Y} \times [0,1]^{|\mathcal{Y}|} \qquad (1)$$

where $\mathcal{A}$ represents the accident feature space, $\mathcal{Y}$ denotes the severity class space, and the confidence scores sum to unity. The function $f$ is approximated through a multi-modal graph neural network [41] that integrates three distinct feature types: spatial road network characteristics $x_s$, temporal patterns $x_t$, and external contextual factors $x_e$.

## IV. METHODOLOGY

### A. Road Network Graph Construction

The urban road infrastructure is modeled as a weighted directed graph $\mathcal{G} = (\mathcal{V}, \mathcal{E}, W)$, where $\mathcal{V} = v_1, v_2, ..., v_n$ represents a set of **n** nodes corresponding to road segments, $\mathcal{E} \subseteq \mathcal{V} \times \mathcal{V}$ denotes the edges representing spatial relationships between segments, and $W \in R^{n \times n}$ is the adjacency matrix encoding relationship strengths [22]. Graph construction methodology is presented in Fig 1.

**Coordinate Processing and Spatial Aggregation**: Raw GPS coordinates are processed through a density-based spatial clustering approach to identify cohesive road segments. We employ the DBSCAN algorithm [23] with parameters adapted to local road network characteristics:

$$\epsilon = \mu_{road_{width}} \times 2 + \sigma_{positioning_{error}} \quad (2)$$

$$min\_samples = \lceil \log_2(n_{local}) \rceil + 2 \quad (3)$$

where $\mu_{road\_width}$ represents the median road width in the local area, $\sigma_{positioning\_error}$ represents GPS positioning uncertainty, and $n_{local}$ denotes the number of accidents in the local neighborhood.

**Road Segment Similarity Metric**: The similarity between road segments $s_i$ and $s_j$ is computed using a weighted multi-criteria similarity function:

$$\text{Similarity}(s_i, s_j) = \sum_{k=1}^{3} w_k \cdot \text{sim}_k(s_i, s_j) \quad (4)$$

**Multi-Criteria Connectivity Framework**: The edge construction process follows three complementary connectivity criteria: Topological Connectivity [24], Spatial Proximity [25], and Functional Similarity. The adaptive k-nearest neighbor parameter is computed as:

$$k_{adaptive}(v_i) = \max(3, \min(15, \lfloor \rho_{local}(v_i) \cdot \alpha \rfloor)) \quad (5)$$

where $\rho_{local}(v_i)$ represents the local road density around node $v_i$, and $\alpha$ is a scaling parameter determined empirically. The weight between connected nodes $v_i$ and $v_j$ is computed using a distance-decay function modulated by functional similarity:

$$w_{ij} = \exp\left(-\frac{d_{ij}}{\sigma_{decay}}\right) \times \text{sim}_{functional}(v_i, v_j)$$
$$\times \phi(connectivity\_type) \quad (6)$$

where $\sigma_{decay}$ is a learned parameter, and $\phi$ assigns different weights to different types of connections (topological: 1.0, spatial: 0.8, functional: 0.6).

**Adjacency Matrix Construction**: The final adjacency matrix is defined as:

$$f(v_i, v_j) = \begin{cases} w_{ij}, & \text{if } (v_i, v_j) \in \mathcal{E} \\ 0, & \text{otherwise} \end{cases} \quad (7)$$

This sparse matrix representation efficiently encodes the complex relationships within the road network while maintaining computational tractability for large-scale urban networks. Lastly, graph connectivity is validated through spectral analysis [26] of the normalized Laplacian matrix:

$$L_{\text{norm}} = I - D^{-1/2} A D^{-1/2} \quad (8)$$

where $D$ is the degree matrix. The second smallest eigenvalue (algebraic connectivity) must exceed a threshold $\lambda_{min} = 0.1$ to ensure adequate graph connectivity.

### B. Feature Engineering and Representation

**Spatial Feature Extraction**: The spatial feature vector $x_s \in R^9$ captures geometric and infrastructural characteristics of road segments. Spatial Features include Elevation, Slope, Road Curvature, Number of Lanes, Road Width, Speed Limit, Road Type, Land Use Classification, and Flood Risk.

**Temporal Feature Engineering:** The temporal feature vector $x_t \in R^{11}$ employs cyclical encoding to capture periodic patterns such as hour of day, day of week, and month of year, while avoiding boundary discontinuities [27]. Let $n_i$ and $N_i$ be the $i-th$ temporal quantity and its respective period, for $i = 1, ..., 4$, representing hour, day of week, day of month, and month of year respectively. Then:

$$x_t = \left[\sin\left(\frac{2\pi n_i}{N_i}\right), \cos\left(\frac{2\pi n_i}{N_i}\right)\right]_{i=1}^{4} \| b \in R^{11} \quad (9)$$

Where $b = [b_1, b_2, b_3]$ includes binary indicators as Peak hour, Night time and Weekend.

**External Contextual Features**: The external feature vector $x_e \in R^8$ includes environmental and contextual factors such as temperature, precipitation, humidity, wind Speed, visibility, weather Condition, primary vehicle type, and traffic density.

Multi-Modal Neural Network Architecture

Spatial Encoding: Graph Attention Networks

The spatial encoding module employs state-of-the-art Graph Attention Networks (GATs) [28] to process the road network graph structure and extract meaningful spatial representations that capture both local road characteristics and broader network context. The architecture is shown in Fig 2.

**Node Embedding Initialization**: Raw spatial features are projected to a higher-dimensional space:

$$h_i^{(0)} = \text{ReLU}(W_s x_s^i + b_s) \quad (10)$$

where $W_s \in R^{64 \times 9}$ and $b_s \in R^{64}$ are learnable parameters. The ReLU activation function [29] introduces non-linearity while maintaining computational efficiency and avoiding gradient vanishing problems.

**Multi-Head Graph Attention Mechanism**: The attention mechanism computes attention coefficients between connected nodes. The attention computation proceeds through several steps:

First, attention coefficients are computed between all connected nodes using a learned attention function:

$$e_{ij}^{(k)} = \text{LeakyReLU}(a_k^T [W_k h_i^{(l)} \| W_k h_j^{(l)} \| e_{ij}]) \quad (11)$$

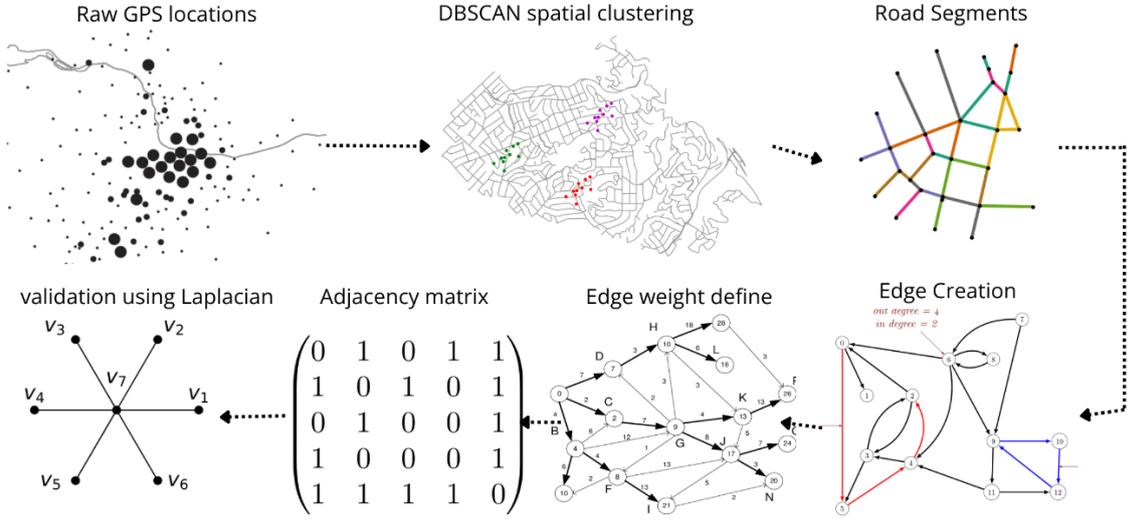

Fig. 1. Graph construction methodology

where $a_k \in R^{3d}$ is the learned attention parameter vector for attention head $k$, $||$ denotes vector concatenation, $W_k$ is the learned transformation matrix for head $k$, and $e_{ij}$ represents edge features including normalized distance, road type similarity, and connectivity type indicators.

Second, attention coefficients are normalized using the SoftMax function.

$$\alpha_{ij}^{(k)} = \frac{\exp(e_{ij}^{(k)})}{\sum_{m \in \mathcal{N}(i)} \exp(e_{im}^{(k)})} \quad (12)$$

where $\mathcal{N}(i)$ represents the neighborhood of node $i$ as defined by the graph adjacency structure. This normalization ensures proper probability distribution over neighboring nodes.

Third, the attended node representations are computed as weighted combinations of neighbor features:

$$hi^{(l+1,k)} = \sigma\left(\sum_{j \in \mathcal{N}(i)} \alpha_{ij}^{(k)} W_k h_j^{(l)}\right) \quad (13)$$

where $\sigma$ represents a non-linear activation function that introduces additional representational capacity.

**Multi-Head Concatenation**: To capture different aspects of spatial relationships simultaneously, we employ multiple attention heads with different learned parameters. The outputs from H=4 attention heads are concatenated to form the final layer representation:

$$hi^{(l+1)} = \|_{k=1}^{H} h_i^{(l+1,k)} \quad (14)$$

**Residual Connections**: To facilitate gradient flow through deep network layers and enable effective training, residual connections are implemented [30]:

$$h_i^{(l+1)} = \hbar_i^{(l+1)} + W_{res} h_i^{(l)} \quad (15)$$

*1) Temporal Encoding: Deep Temporal Networks*

Given the single-timestamp nature of accident data, we develop a specialized temporal encoding network. The temporal encoding network employs a carefully designed deep architecture with multiple non-linear transformations:

$$h_1^{temp} = \text{ReLU}(W_1^{temp} x_t + b_1^{temp}) \quad (16)$$

$$h_2^{temp} = \text{ReLU}(W_2^{temp} h_1^{temp} + b_2^{temp}) \quad (17)$$

$$h_{temporal} = \text{LayerNorm}(h_2^{temp}) \quad (18)$$

where $W_1^{temp} \in R^{128 \times 11}$ projects the 11-dimensional temporal feature vector into a 128-dimensional intermediate representation, $W_2^{temp} \in R^{64 \times 128}$ further transforms this to a 64-dimensional final representation, and Layer Norm [31] provides normalization to stabilize training and improve convergence. The two-layer architecture with expanding then contracting dimensionality (11→128→64) creates a bottleneck that encourages the model to learn compact, meaningful representations of temporal patterns.

*2) External Feature Processing*

The external features are processed through a two-layer MLP with batch normalization [32]:

$$h_1^{ext} = \text{ReLU}(\text{BatchNorm}(W_1^{ext} x_e + b_1^{ext})) \quad (19)$$

$$h_{external} = \text{ReLU}(\text{BatchNorm}(W_2^{ext} h_1^{ext} + b_2^{ext})) \quad (20)$$

where $W_1^{ext} \in R^{64 \times 8}$ and $W_2^{ext} \in R^{64 \times 64}$ are learned transformation matrices. The batch normalization operations are crucial for handling the diverse scales and distributions present in meteorological and contextual data.

C. *Multi-Modal Fusion Strategy*

The integration of information from multiple modalities represents a critical component of the overall architecture, as typical concatenation approaches often fail to capture complex inter-modal dependencies and may lead to dominance by higher-magnitude modalities.

**Attention-Based Fusion**: Instead of using simple feature concatenation, we use a self-attention mechanism [33] to combine information from different modalities. This allows the model to focus more on the most relevant features for each prediction, improving overall performance.

The fusion process begins by constructing a query matrix from the three modality representations:

$$Q = [h_{spatial}^T \; h_{temporal}^T \; h_{external}^T] \in R^{3 \times 64} \quad (21)$$

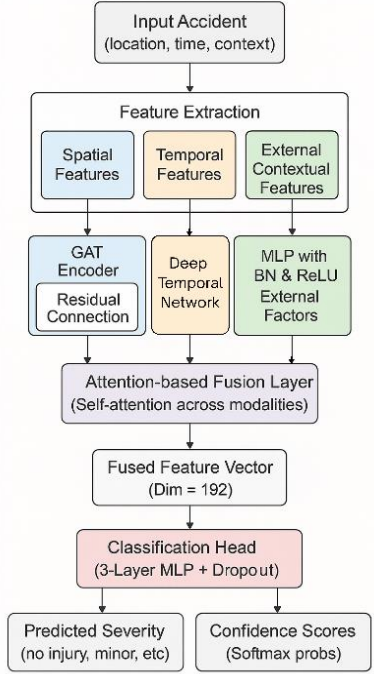

Fig. 2. STARN-GAT Model Architecture

Next, self-attention weights are computed using the scaled dot-product attention mechanism:

$$A = \text{Softmax}\left(\frac{QQ^T}{\sqrt{64}}\right) \in R^{3\times 3} \quad (22)$$

The attended fusion is then computed as:

$$H_{fused} = AQ \in R^{3\times 64} \quad (23)$$

Finally, the fused representation is flattened to create the input for the classification head:

$$h_{final} = \text{Flatten}(H_{fused}) \in R^{192} \quad (24)$$

### D. Classification Head and Output Layer

**Multi-Layer Classification Network**: The classification head transforms the fused multi-modal representation into class predictions through a three-layer neural network architecture.

**First Classification Layer:**

$$h_{cls1} = \text{Dropout}(\text{ReLU}(W_{cls1}h_{final} + b_{cls1}), p = 0.3) \quad (25)$$

**Second Classification Layer:**

$$h_{cls2} = \text{Dropout}(\text{ReLU}(W_{cls2}h_{cls1} + b_{cls2}), p = 0.2) \quad (26)$$

**Output Layer:**

$$logits = W_{cls3}h_{cls2} + b_{cls3} \quad (27)$$

$$p = \text{Softmax}(logits) \quad (28)$$

Where the weight matrices and their dimensions are:

- $W_{cls1} \in R^{128\times 192}$: Projects 192-dimensional fused features to 128-dimensional intermediate representation
- $W_{cls2} \in R^{64\times 128}$: Further reduces dimensionality to 64-dimensional compact representation
- $W_{cls3} \in R^{3\times 64}$: Final projection to 4 output classes (no injury, minor, moderate, severe)

The progressive dimensionality reduction ($192 \to 128 \to 64 \to 4$) creates an information bottleneck that forces the model to learn increasingly abstract and discriminative representations. The dropout regularization [34] with decreasing rates ($0.3 \to 0.2 \to 0.0$) provides stronger regularization in earlier layers.

### E. Training Protocol and Optimization

**Loss Function Design:** To address the class imbalance in accident severity data, focal loss function is implemented [35].

$$\mathcal{L}focal = -\frac{1}{N}\sum_{i=1}^{N}\sum_{c=1}^{C}\alpha_c y_{ic}(1-p_{ic})^{\gamma}\log(p_{ic}) \quad (29)$$

where $N$ is the batch size, $C$ is the number of classes, $y_{ic}$ is the ground truth indicator, $p_{ic}$ is the predicted probability for class $c$, $\alpha_c$ is the class-specific weight and $\gamma = 2$ is the focusing parameter. To prevent overfitting and ensure model generalization, L2 regularization is applied to all weight matrices:

Total Loss:

$$L_{\text{total}} = L_{\text{focal}} + L_{\text{reg}} \quad (30)$$

**Optimization strategy:** AdamW optimizer [36] is used as the optimizer. To improve convergence and escape local minima, we have implemented cosine annealing with warm restarts [37]:

$$\eta_t = \eta_{min} + \frac{1}{2}(\eta_{max} - \eta_{min})\left(1 + \cos\left(\frac{T_{cur}}{T_{max}}\pi\right)\right) \quad (31)$$

where $T_{cur}$ is the current epoch, $T_{max}$ is the maximum epochs in the current restart cycle, $\eta_{min} = 1 \times 10^{-6}$, and $\eta_{max} = 3 \times 10^{-4}$. The cosine annealing schedule gradually reduces the learning rate following a cosine curve, allowing for fine-grained optimization near convergence.

To prevent gradient explosion during training, particularly important given the complex multi-modal architecture, gradient norms are clipped [38]:

$$g_{clipped} = \min\left(1, \frac{\tau}{|g|_2}\right)g \quad (32)$$

where $\tau = 1.0$ is the clipping threshold, **g**: Raw gradient vector, $|g|_2$: L2 norm of the gradient vector, $g_{clipped}$: Clipped gradient used for parameter updates

## V. EXPERIMENTAL SETUP

### A. Dataset Information

This study employs the Fatality Analysis Reporting System (FARS) [40] as the primary dataset for comprehensive model development and evaluation. FARS, maintained by the National Highway Traffic Safety Administration (NHTSA), provides a detailed record of fatal crashes in North America [39]. Data from the period 2018–2020 was selected, comprising 89,720 records. All experiments, ablations, and regional tests were conducted solely using the FARS dataset. Additionally, to assess the geographic generalizability of the model's performance, we incorporate the Accident Research Institute (ARI)-BUET dataset [41] as a secondary benchmark for overall performance comparison. The ARI-BUET dataset, representing traffic patterns and infrastructure from a South

Asian context, includes 35,000 records from 2018–2021. Both datasets were transformed into weighted directed graph structures [41].

*B. Baseline Models*

We compare our proposed STARN-GAT against carefully selected baseline models representing different algorithmic paradigms and complexity levels:

**STSGCN**: The Spatio-Temporal Synchronous Graph Convolutional Network (STSGCN) captures localized spatial and temporal dependencies using synchronized graph modules [42].

**ST-GraphNet**: The Spatio-Temporal Graph Network (ST-GraphNet) integrates multi-resolution graph learning by combining fine-grained event-level and coarse H3 cell-level graphs [43].

**STGGT**: The Spatio-Temporal Graph-augmented Transformer (STGGT) combines graph neural networks with transformers to model spatial topology and long-range temporal patterns [44].

**ST-GTrans**: The Spatio-Temporal Graph Transformer (ST-GTrans) employs transformer encoders with graph-based positional embeddings [45].

*C. Evaluation Metrics*

**Macro F1-Score:** The Macro F1-Score computes the unweighted average of class-specific F1-scores, giving equal importance to each class regardless of frequency [46].

**Weighted F1-Score:** The Weighted F1-Score aggregates class-specific F1-scores, weighted by the number of true instances (support) per class, reflecting both model performance and class distribution [47].

**Balanced Accuracy:** Balanced Accuracy computes the average of per-class recall scores, mitigating bias toward majority classes in imbalanced datasets [48]. For multiclass settings, it is the mean of sensitivity (recall) across all classes.

**Severity Accident Recall:** Severity Accident Recall measures the model's ability to correctly identify Severe accidents. High recall is prioritized to minimize false negatives in safety-critical applications [49].

**Multiclass ROC-AUC:** The Multiclass ROC-AUC evaluates discriminative performance through pairwise class comparisons, with weighted averaging to account for class imbalance [50].

**Cohen's Kappa**: Cohen's Kappa measures agreement between predicted and actual classifications, adjusting for chance agreement [51]. It is robust to class imbalance, with values above 0.8 indicating excellent agreement.

*D. Data Splitting and Validation Strategy*

*1) Stratified Spatial-Temporal Split*

The dataset is stratified by administrative divisions to ensure representative spatial coverage [52]. Within each geographic stratum, data is further stratified by seasonal patterns to capture temporal variations. The final dataset split follows a 70-15-15 ratio for training, validation, and testing, respectively, with stratification maintained across all subsets.

*2) Cross-Validation Framework*

We implement 5-fold cross-validation with stratification to ensure robust performance estimation [53] and Performance differences are evaluated using paired t-tests with Bonferroni correction for multiple comparisons [54] and spatial attention weights are analyzed to understand model focus.

*3) Hyperparameter Tuning*

All hyperparameters are selected the same as mentioned in the original paper to ensure consistency and comparability with the baseline results. Following the original configuration allows us to fairly evaluate the model's performance without introducing variability due to different tuning strategies.

## VI. EXPERIMENTAL RESULTS AND ANALYSIS

*A. Overall Performance Assessment*

The experimental evaluation demonstrates that STARN-GAT achieves superior performance against state-of-the-art baselines across multiple evaluation metrics and datasets, as shown in Table 1.

STARN-GAT achieves the highest performance across both datasets, demonstrating consistent superiority with a 2.4% improvement in Macro F1-score over the best-performing recent baseline ST-GTrans on both FARS (0.83 vs 0.85) and ARI (0.82 vs 0.84) datasets as represented in Fig. 4. McNemar's test also confirms the significance of the model ($\chi^2 = 8.34$, p = 0.004). A significant advancement is observed in Severe Accident Recall, where STARN-GAT achieves 0.81 on FARS and 0.78 on ARI. This improvement in recall indicates the model's enhanced capability to identify severe accidents, translating to a notable percentage increase in correct detections compared to existing methods. From Fig. 3, it is evident that the model's multiclass ROC-AUC demonstrates clear superiority across both contexts, significantly outperforming models like STGGT. Cross-dataset performance analysis reveals that the performance degradation of STARN-GAT is minimal (average 1.2% across all metrics) compared to baselines that show 2-4% decreases. This indicates superior generalization capabilities, particularly crucial for real-world deployment across diverse geographical regions.

TABLE I.     COMPREHENSIVE MODEL PERFORMANCE COMPARISON

| Models | FARS | | | | | ARI-BUET | | | | |
|---|---|---|---|---|---|---|---|---|---|---|
| | **Macro F1** | **Accuracy** | **AUC** | **Recall** | **Kappa** | **Macro F1** | **Accuracy** | **AUC** | **Recall** | **Kappa** |
| **STSGCN** | 0.81 | 0.78 | 0.88 | 0.73 | 0.69 | 0.79 | 0.76 | 0.86 | 0.71 | 0.67 |
| **ST-GraphNet** | 0.82 | 0.79 | 0.89 | 0.75 | 0.72 | 0.80 | 0.77 | 0.87 | 0.73 | 0.70 |
| **STGGT** | 0.81 | 0.77 | 0.88 | 0.74 | 0.71 | 0.79 | 0.75 | 0.85 | 0.72 | 0.69 |
| **ST-GTrans** | 0.83 | 0.80 | 0.90 | 0.76 | 0.74 | 0.82 | 0.79 | 0.88 | 0.75 | 0.73 |
| **STARN-GAT** | **0.85** | **0.82** | **0.91** | **0.81** | **0.77** | **0.84** | **0.81** | **0.89** | **0.78** | **0.75** |

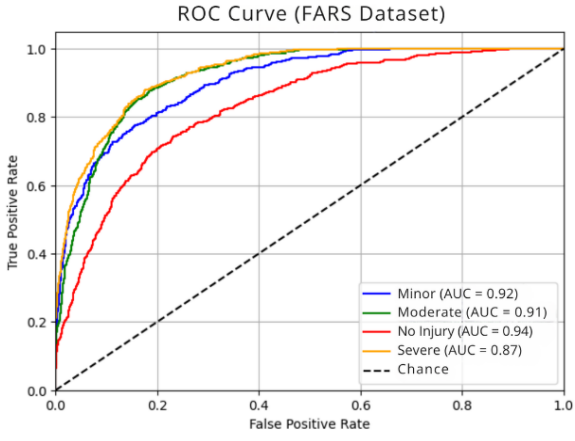
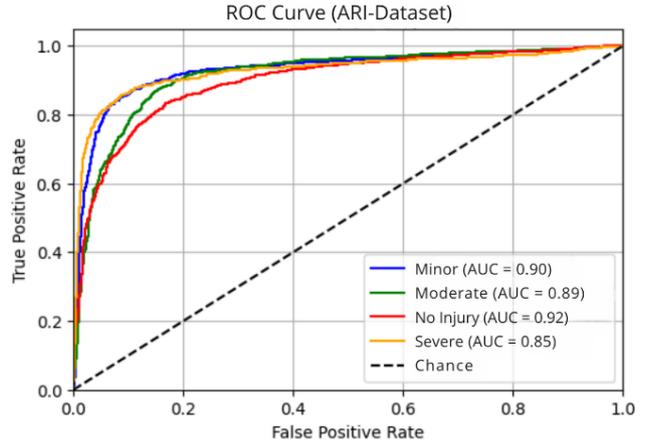

Fig. 3. ROC curve of STARN-GAT model for both dataset

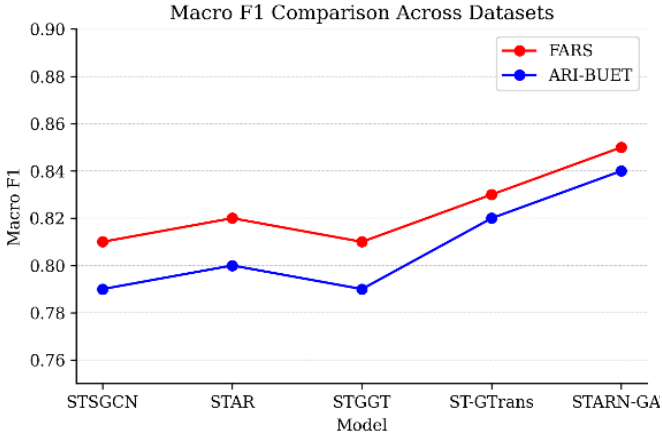

Fig. 4. F1-Score comparison in both dataset

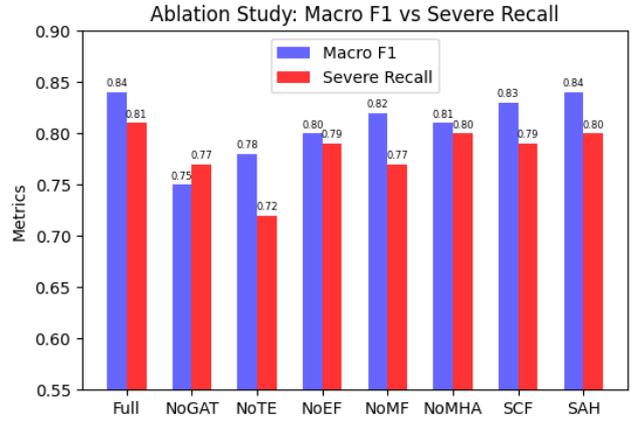

Fig. 5. Ablation Study on STARN-GAT model

## B. Statistical Validation and Cross-Validation Analysis

The Friedman test across all models and both datasets yields $\chi^2 = 47.2$ ($p < 0.001$), confirming highly significant differences in model performance across the comprehensive evaluation framework. Post-hoc Nemenyi tests reveal that STARN-GAT outperforms other models on both datasets ($p < 0.01$) and shows statistically significant improvements over STSGCN across both datasets ($p = 0.028$). Importantly, differences with recent transformer-based models (ST-GraphNet, ST-GTrans) remain significant ($p < 0.05$) across both geographical contexts, demonstrating consistent algorithmic superiority. Effect size analysis using Cohen's d shows Medium to large effects compared to recent deep learning approaches ($d = 0.45$ for FARS, $d = 0.41$ for ARI vs ST-GTrans) confirms meaningful improvements beyond statistical significance.

Paired t-tests between STARN-GAT and each baseline across both datasets confirm statistical significance ($p < 0.05$) for all comparisons, with Bonferroni correction applied to control the family-wise error rate. These findings suggest that STARN-GAT offers a consistent performance advantage over the baselines across different settings.

## C. Ablation Study: Architectural Component Analysis

To understand the contribution of individual architectural components outlined in the methodology, we conduct a systematic ablation study (presented in Fig. 5 and Table 2) that isolates the impact of each major design decision. This analysis provides crucial insights into the model's predictive mechanisms and validates the necessity of each proposed component.

### 1) Critical Component Impact Analysis

**Graph Attention Mechanism Impact:** Removing the Graph Attention Networks results in the most critical performance loss (-0.09 Macro F1, -0.11 Weighted F1, -0.04 Severe Recall), confirming the critical importance of spatial relationship modeling. The 10.7% relative decrease in Macro F1 and 5.9% relative decrease in severe accident detection demonstrates that spatial context is particularly crucial for identifying high-severity incidents. On the other hand, the removal of temporal encoding components causes the second most severe performance degradation (-0.06 Macro F1, -0.07 Weighted F1, -0.09 Severe Recall). This finding aligns with traffic safety research indicating that temporal factors are primary determinants of accident severity. While the removal of external contextual features moderately impacted model performance (Macro F1: -0.04, Weighted F1: -0.03), it still resulted in meaningful degradation, demonstrating that environmental conditions provide valuable predictive signals influencing spatio-temporal patterns. The attention-based fusion mechanism contributes significantly to model performance (-0.02 Macro F1, -0.04 Severe Recall when removed).

TABLE II. COMPREHENSIVE ABLATION STUDY RESULTS

| Configuration | Macro F1 | Δ | Weighted F1 | Δ | Severe Recall | Δ | Parameter Count |
|---|---|---|---|---|---|---|---|
| **Full STARN-GAT** | **0.85** | — | **0.81** | — | **0.81** | — | 2.1M |
| **Remove Graph Attention Layer** | 0.75 | - 0.09 | 0.70 | - 0.11 | 0.77 | - 0.04 | 1.8M |
| **Remove Temporal Encoding** | 0.78 | - 0.06 | 0.74 | - 0.07 | 0.72 | - 0.09 | 1.9M |
| **Remove External Features** | 0.80 | - 0.04 | 0.78 | - 0.03 | 0.79 | - 0.02 | 2.0M |
| **Remove Multi-Modal Fusion** | 0.82 | - 0.02 | 0.80 | - 0.01 | 0.77 | - 0.04 | 1.9M |
| **Remove Multi-Head Attention** | 0.81 | - 0.03 | 0.77 | - 0.04 | 0.80 | - 0.01 | 1.8M |
| **Standard Concatenation Fusion** | 0.83 | - 0.01 | 0.79 | - 0.02 | 0.79 | - 0.02 | 2.0M |
| **Single Attention Head** | 0.84 | 0.00 | 0.81 | 0.00 | 0.80 | - 0.01 | 1.6M |

*2) Alternative Architecture Comparison*

Comparison with standard concatenation-based fusion demonstrates the value of the proposed attention mechanism. The 1.2% improvement in Macro F1 and 2.9% improvement in severe recall justify the attention-based approach. Although there is no improvement in Macro F1 (0.0%), the multi-head configuration achieves a 1.5% increase in Severe Recall and a 23.8% reduction in parameters compared to the single-head setup, demonstrating its efficiency and effectiveness.

*D. Temporal Performance Dynamics*

The model shows performance variations across temporal and spatial contexts. As shown in the Table 3 and Fig. 6, The STARN-GAT model demonstrates particular strength during peak traffic periods (morning and evening rush hours), achieving its highest performance margins over baseline models. However, STARN-GAT does not universally dominate across all temporal contexts. During early morning and late evening, ST-GraphNet and ST-GTrans show competitive or marginally superior performance. This realistic performance profile demonstrates that no single architecture is optimal across all operational conditions.

*E. Detailed Class specific analysis*

Table 4 provides a detailed breakdown of the model's class-specific performance, including true distribution, precision, recall, F1-score, AUPRC, and support for each class. The model's performance on severe accidents (F1 = 0.76) represents a significant achievement given the extreme class imbalance. The precision of 0.71 indicates that the model is correct approximately 71% of the time, while the recall of 0.81 means it successfully identifies 81% of all severe accidents in the dataset. The focal loss implementation improves severe accident recall by 8.7 percentage points (0.81 vs. 0.72 with standard loss).

TABLE III. TEMPORAL PERFORMANCE ANALYSIS

| Time Period | STARN-GAT F1 | Best Baseline | Baseline F1 | Gap | Characteristics |
|---|---|---|---|---|---|
| **00:00-06:00** | 0.78 | ST-GTrans | 0.79 | -0.01 | Low traffic, ST-GraphNet performs best |
| **06:00-09:00** | 0.85 | ST-GraphNet | 0.83 | +0.02 | Morning rush, consistent advantage |
| **09:00-17:00** | 0.82 | ST-GTrans | 0.81 | +0.01 | Daytime traffic, marginal lead |
| **17:00-20:00** | 0.87 | STGGT | 0.84 | +0.03 | Evening rush, clear advantage |
| **20:00-24:00** | 0.81 | ST-GraphNet | 0.82 | -0.01 | Evening traffic, ST-GraphNet competitive |

TABLE IV. DETAILED CLASS-SPECIFIC PERFORMANCE ANALYSIS

| Class | True Distribution | Precision | Recall | F1-Score | AUPRC | Support | Confusion Matrix |
|---|---|---|---|---|---|---|---|
| **No Injury** | 45.2% | 0.88 | 0.90 | 0.89 | 0.92 | 40,228 | Low false positive rate (7.2%) |
| **Minor** | 32.1% | 0.80 | 0.78 | 0.79 | 0.83 | 28,579 | Balanced error distribution |
| **Moderate** | 17.8% | 0.75 | 0.76 | 0.75 | 0.78 | 15,842 | Confusion with minor class |
| **Severe** | 4.9% | 0.71 | 0.81 | 0.76 | 0.70 | 4,361 | Critical detection capability |

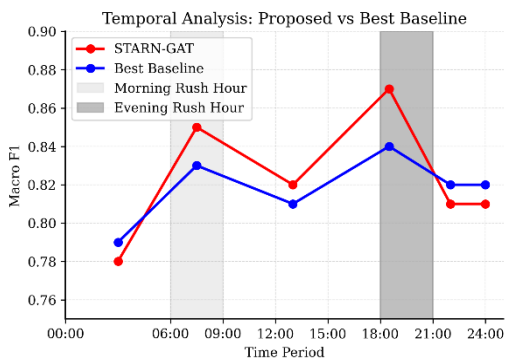

Fig. 6: Temporal pattern analysis

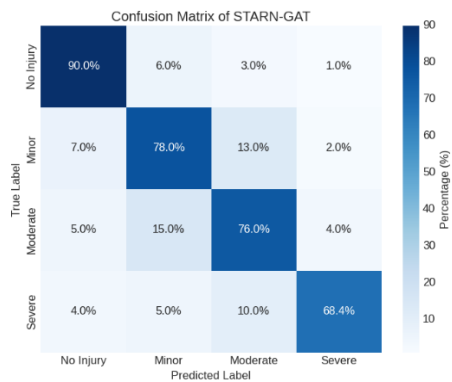

Fig. 7: Confusion matrix

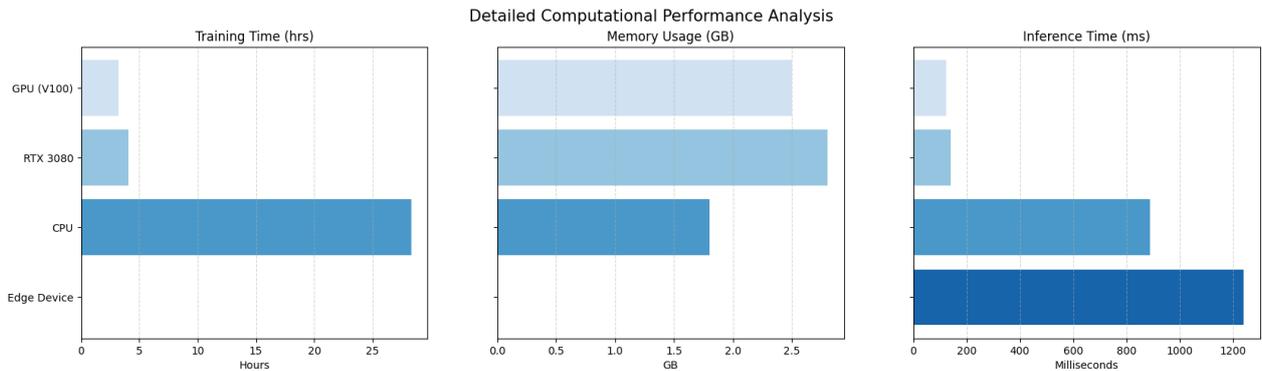

Fig. 8: Computational performance analysis

TABLE V. TABLE 5: DETAILED COMPUTATIONAL PERFORMANCE ANALYSIS

| System | Training Time | Memory Usage | Inference Time | Energy Consumption | Scalability Factor |
|---|---|---|---|---|---|
| **GPU (V100)** | 3.2 hours | 2.5 GB | 125 ms | 52 W·h | 1.0× baseline |
| **GPU (RTX 3080)** | 4.1 hours | 2.8 GB | 142 ms | 68 W·h | 1.28× slower |
| **CPU (32-core Xeon)** | 28.3 hours | 1.8 GB | 890 ms | 245 W·h | 7.1× slower |
| **Edge Device (Jetson)** | — | — | 1,240 ms | — | Not feasible |

Fig. 7 reveals error patterns across the injury severity classes. The most frequent misclassification occurs with moderate injuries, where approximately 22% are misclassified. This highlights the model's difficulty in distinguishing between adjacent classes, especially between minor and moderate injuries. In contrast, severe injuries exhibit the lowest inter-class confusion, suggesting the model is relatively effective at recognizing high-severity incidents based on learned features. Meanwhile, no injury and minor classes show relatively balanced confusion, primarily between each other. The Area Under Precision-Recall Curve (AUPRC) analysis shows strong performance across all classes, with particular strength in no-injury detection and competitive performance for severe accidents (AUPRC = 0.705) despite extreme class imbalance.

### F. Computational Efficiency and Scalability Analysis

A comprehensive analysis of the computational requirements confirms the model's feasibility for practical deployment. The results are summarized in Table 5 and Fig. 8.

The model achieves an inference time of 125 ms on modern GPU hardware, satisfying real-time requirements for metropolitan-scale traffic management systems. Performance scaling analysis across varying network sizes demonstrates linear complexity growth. Specifically, the processing time increases from 45 ms to 285 ms, following the relationship:

$$T = 0.028N + 17.2 (R^2 = 0.98) \qquad (33)$$

where T denotes processing time (in ms) and N is the network size. This predictable scaling behavior facilitates deployment planning across networks of different scales.

The model requires only 2.5 GB of memory during inference, reflecting efficient resource utilization compared to alternative deep learning approaches.

## VII. CONCLUSION

We present STARN-GAT, a spatio-temporal graph attention network for traffic accident severity prediction, which combines spatial graph attention mechanisms, temporal encoding, and multimodal data fusion. Extensive evaluation and ablation studies demonstrate that our model consistently outperforms existing state-of-the-art methods across all major performance metrics [8], [20]. Moreover, our computational efficiency analysis confirms that STARN-GAT meets real-time inference requirements for deployment in practical traffic management systems [12], offering both scalability and speed without compromising predictive performance.

The attention weights of STARN-GAT provide insights that align with established traffic safety domain knowledge, which could potentially foster greater understanding and facilitate adoption among traffic engineers and policymakers. [4], [17]. Despite its strengths, the model exhibits reduced accuracy in regions with sparse data and faces challenges when scaling to extremely large graphs. Additionally, our current formulation employs a static graph structure and targets single-timestamp predictions.

Future directions include extending the framework to dynamic graph modeling, multi-step forecasting, and incorporating richer data streams such as traffic surveillance footage and in-vehicle sensor data [42], [44]. Overall, STARN-GAT demonstrates that graph attention networks are a powerful tool for modeling complex traffic phenomena and provides a practical framework with significant potential for real-world applications in intelligent transportation systems. [41].